\documentclass{article}

\usepackage{amsmath}
\usepackage{caption}
\usepackage{arxiv}
\usepackage{graphicx}
\usepackage[utf8]{inputenc} % allow utf-8 input
\usepackage[T1]{fontenc}    % use 8-bit T1 fonts
\usepackage{hyperref}       % hyperlinks
\usepackage{url}            % simple URL typesetting
\usepackage{booktabs}       % professional-quality tables
\usepackage{amsfonts}       % blackboard math symbols
\usepackage{nicefrac}       % compact symbols for 1/2, etc.
\usepackage{microtype}      % microtypography
\usepackage{lipsum}
\usepackage{multicol}
\usepackage{multirow}
\usepackage{subcaption}
\usepackage{arxiv}
\usepackage[square, numbers, comma, sort&compress]{natbib}

\title{3D tissue reconstruction with Kinect to evaluate neck lymphedema}

\author{Gerrit Brugman, Beril Sirmacek \\
  Robotics and Mechatronics\\
  University of Twente\\
  Enschede, The Netherlands \\
  \texttt{b.sirmacek@utwente.nl} \\
}

\begin{document}
\maketitle

\begin{abstract}
Lymphedema is a condition of localized tissue swelling caused by a damaged lymphatic system. Therapy to these tissues is applied manually. Some of the methods are lymph drainage, compression therapy or bandaging. However, the therapy methods are still insufficiently evaluated. Especially, because of not having a reliable method to measure the change of such a soft and flexible tissue. In this research, our goal has been providing a 3d computer vision based method to measure the changes of the neck tissues. To do so, we used Kinect as a depth sensor and built our algorithms for the point cloud data acquired from this sensor. The resulting 3D models of the patient necks are used for comparing the models in time and measuring the volumetric changes accurately. Our discussions with the medical doctors validate that, when used in practice this approach would be able to give better indication on which therapy method is helping and how the tissue is changing in time.
\end{abstract}

% keywords can be removed

\section{Introduction and Background}

According to NCI (2017), neck lymphedema is a side effect of the removal of lymph nodes due to cancer. If the lymph nodes are removed, the flow of lymph may be slower and the lymph could collect in the tissues, causing swelling. Unfortunately, to the best of our knowledge based on the literature search and doctor discussions, there is no highly reliable instrument to measure the degree of change in the neck volume. The current method consists of measuring the circumference of the neck which gives an indication of the volume in the neck. This procedure is done by a doctor with a measuring tape. This procedure is error-prone since it relies on the doctor. What they usually do is have the same doctor do the tape measurement at defined intervals since a different doctor might use a slightly different location which influences the diagnosis. 
\par
For bioimpedance the same as the tape measurement occurs, the doctor has to take the measurement exactly at the same locations every defined interval. The measurement is done with a specified positioning protocol. The setup is outlined according to paper Purcell et al. \cite{Purcell}:

1. Body position on the bed: lying face up, no pillow, crown of head aligned with the bed
edge.

\begin{figure}
  \centering
  \includegraphics[scale=0.3]{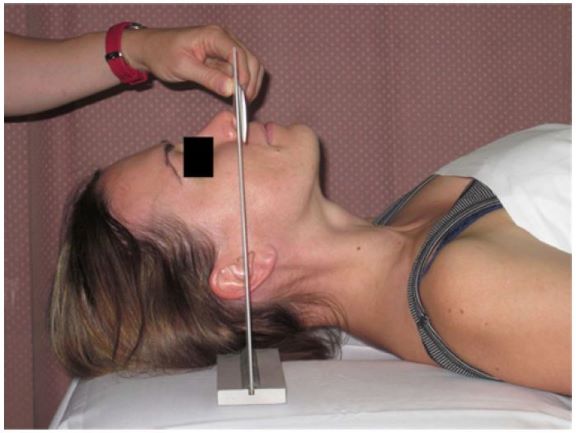}
  \caption{Current measurement techniques (trying to find a good position for the patient head with a vertical ruler) \cite{Purcell}}
  \label{fig:fig1}
\end{figure}

2. Head position: head aligned to form a 90$^{\circ}$ intersection between a ruler and a set square.
Set square: horizontal edge on the bed perpendicular to the subject; 90$^{\circ}$ corner closest to the subject: vertical edge aligned with the inferior margin of the tragus. Ruler: place the ruler at the patient’s inferior nose resting against the base of the spine of the nose. This position can be seen in Figure \ref{fig:fig1}.

The measurement tape was used to get the following measurements:

1. Lower neck circumference

2. Upper neck circumference

3. Length from ear to ear

4. Length from lower lip edge to lower neck circumference

All these measurements are there to get an indication of the volume of the neck. The goal is to
replace the old method mentioned above and replace it by calculating the volume of the neck
using 3D imaging tools.

The research questions that have focused for answering are listed below:

• How many Kinect frames from different angles need to be taken to create a point cloud that can be used for detecting lymphedema?

• How accurate and precise is this method for detecting change in volume in the neck area caused by lymphedema?

\section{Data analysis}

The problem is that there is no reliable instrument to measure the status of neck lymphedema. The solution will be to calculate the volume using a different way of measuring using the Kinect camera.

\subsection{Investigation of the validity of the research questions in medical domain}
During the progress of the experiments, we have conducted interviews with medical doctors from UMC Utrecht. These meetings have validated that computer vision based 3D volume measurement techniques are necessary, especially for observing the volume change of the neck area. Currently, the medical doctors are often using the water-based measurements (putting arms or legs into a bucket and observing the water volume change). Obviously, this method is not applicable for the neck and head area of the patients.

\subsection{Investigation of the validity of the sensor choice}

The Kinect is a gaming accessory made by Microsoft made for the Xbox 360 platform. There are two versions of the Kinect, Kinect v1 and Kinect v2. The first version of the Kinect uses infrared light pattern projections, while the second version uses a time of flight method according to \cite{Immotionar}. The Kinect v2 projects infrared light and on return determines the phase shift to calculate how long the light beam has been traveling. The Kinect v2 is the one used for this project. Besides the Kinect there are other methods of obtaining depth information, namely:

• Real Sense by Intel \cite{Intel}, this is a depth sensor similar to the Kinect. It can be extended to also estimate its position using an accelerometer and gyroscope. This can be useful when reconstructing a full 3D model by moving the camera.

• Stereo vision, this method uses two cameras and by finding the same points in both the images the depth can be estimated using trigonometry.
The Kinect was used above Real Sense because the Kinect was already available, and since both almost behave the same this is not an issue. Stereo vision is very susceptible to surrounding conditions because it has to find the same point in both pictures. Creating a good and reliable depth image using stereo vision is already worthy of creating a new research study about. 

Currently some researchers have already conducted relevant scientific work with in this field using Kinect, below a couple of them are listed:

• ReconstructMe by Heindl et al. \cite{reconstructme}, this is a software that creates real-time 3D models.
This method is not available for Kinect v2 but only for Kinect v1 and some other imaging tools. ReconstructMe uses the accelerometer that is present on the Kinect v1 to estimate the location of the camera and merge depth images based on this.

• KinectFusion by A. Newcombe et al. \cite{Newcombe1}, this uses Iterative Closest Point (ICP) to merge the depth images in real-time.

These methods are all about constructing a full 3D model of a scene or object. KinectFusion
would be perfect for the application to determine the status neck lymphedema but unfortunately
is not open-source.

\subsection{Choosing a reliable method to merge point clouds}

To create a full 3D model of the neck to calculate the volume, multiple depth images need to be merged. A couple of methods can be used for this:

• Point cloud registration, this was used for KinectFusion where they used ICP to register the point cloud to a reference frame. ICP minimizes the difference between two point clouds by transforming one point cloud iteratively. Once a minimum has been found the transformed point cloud can be merged with the reference point cloud.

• When the camera position is already known compared to the reference frame the transformation
matrix does not have to be found using point cloud registration. The transformation
matrix can simply be calculated and applied to the point cloud to merge it into
the reference frame.
For merging the point clouds the second option has been chosen. When using ICP, or any other
method of registering a point cloud, both point clouds already require to be very similar. In our
case, not more than two point clouds will be used, which means that the point clouds are not
that similar. In our case, the camera position compared to the human body is fixed, namely the
front of the patient and the back.

\subsection{Choosing good viewing angles}

When only one depth image is taken it results in a one angle point cloud. This means that with
this method the volume can not be calculated, but half of the circumference of the neck can be.
Half of the circumference of the neck might still give information about the status of the neck
lymphedema. To get half the circumference of the neck the edge points of the neck first have to
be found to conduct the measurement. To get these points two methods are presented below:

• Take the derivative of each row of the image and find the two peaks in the derivative with
the highest prominence. The prominence of a peak measures how much the peak stands
out due to its height and its location relative to other peaks. Collect all these points and conduct the measurement.

• Use the points shown above to find an average depth of the edges. Then find the two
points that match this depth as closely as possible for each row.
After some testing, the first option gave a decent result but it varied too much, the depth of
each point was inconsistent. That is why the second option was introduced, this option gives
all the points at the same depth and makes the measurement more consistent.

\subsection{Choosing a method to compare 3D reconstructed time series data}

When two depth images are taken and also merged into one 3D point cloud, the measurements
can be conducted. To get the volume of the neck the area of each slice from top to bottom will
first be calculated. Then a choice has to be made on where the neck begins and where it ends,
this can be done by getting some reference points from the image. The reference points that can be used can be shown below:

• In the graph of the area of the patient the minimum can be found, this minimum corresponds
to the minimum area of a layer from the neck of the patient. After this, the begin and
endpoint can be a fixed distance up and down of this minimum point. The advantage of this is that it is a simple method and reliable in a normal situation. The disadvantage is that neck lymphedema can cause swelling exactly at this minimum point which causes the minimum to shift and make the results unreliable.

• Before calculating some more specific points can be found first. For example, the lower
lip can be a beginning point and the collarbones the endpoint. This would be a more
reliable option when the neck starts swelling compared to the previous option. But this
requires finding reliable points in a depth image, the lower lip is doable but the collarbone
is hard to detect.
The first option was chosen because it is a simple solution to test how accurate and precise the
solution is. But when it will be implemented to determine the status of neck lymphedema the
second method has to be implemented.

\subsection{Experiment Design}
In this section, the final design of two different methods will be further elaborated. The first method is using one depth image to conduct measurements on, the second method merges two depth images to conduct measurements on.

\subsubsection{Processing each point cloud input from certain angles}
\begin{figure}
  \centering
  \includegraphics[scale=0.6]{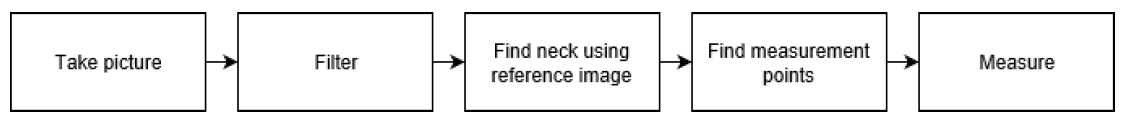}
  \caption{Flowchart of the method using one depth image}
  \label{fig:fig2}
\end{figure}

In this section, a method for conducting measurements on a single angle point cloud will be
explained. It is assumed that one imaging session was taken before, which will be used as the
reference to compare with. This reference depth image is a depth image taken right after cancer
has been removed. This image used for this can be seen in the result section. The flow of this
method can be seen in the flowchart shown in figure \ref{fig:fig2}.

\subsubsection{Position of the patient and the Kinect sensor}
The image will be taken from the front, this is due to lymphedema causing swelling mostly on
the sides and front of the neck. The patient has to be standing in front of the camera at a fixed
distance of one meter from the camera. The setup can be seen in \ref{fig:fig3}.

\subsubsection{Filtering the input point clouds}
The depth image that has been taken has to be filtered to remove false measurements. This
is done by detecting outliers using the function filloutliers() in MATLAB (2019). The function
defines outliers as follows: An outlier is a value that is more than three scaled median absolute
deviations (MAD) away from the median. The MAD is defined as:

\begin{equation}
    MAD = median(|A_i-median(A)|)
\end{equation}

and the scaled MAD: $\sigma$, which is a consistent estimator for the standard deviation, is defined as:
\begin{equation}
    \sigma = k \times MAD
\end{equation}

The function will operate on each column of the image, which is the vertical axis. It will use a moving window for detecting the outliers. Once found it uses linear interpolation, based on
surrounding points, to replace the point. The size of the window can be varied to achieve the best result without any outliers.

\subsection{Finding a good position to start data acquisition}

Before comparing the neck with the reference image, taken in the previous session, the neck
has to be located in the current image. The neck from the reference image can be used as a tool
to find the neck in the current image. This can be realized using the normalized correlation
coefficient. The normalized correlation coefficient can be computed as follows:

\begin{equation}
    c[m,n] = \dfrac{\sum_{k,l}x[k,l]h[m-k,n-l]}{\sqrt{\sum_{k,l}(x[k,l])^2g[m-k,n-l]-\dfrac{1}{KL}(\sum_{k,l}x[k,l]g[m-k,n-l])^2}}
\end{equation}

where x[k,l] is the image, h[m,n] is the neck mirrored on the horizontal and vertical axis from
the reference image. K and L the size of h and g[m,n] is a impulse response with same size
as h but filled with only ones. This g[m,n] will sum all the pixels in the sliding window. The
normalized correlation factor will have values between -1 and 1, the larger the value the larger
the similarity. After this, the pixel with the highest correlation factor can be found. Once it is
found the neck image can be used for further measurements.

\subsection{Measuring the change in 3D reconstruction time series}
As mentioned in the analysis the method of getting the start and endpoints to measure half of
the circumference was achieved by first taking the derivative of each row and finding the two
peaks with the highest peak prominence. After these peaks have been found for each row in the depth image, the average depth of all these points will be calculated. When this depth is known,
the two points that are closest to this value will be used as begin and endpoints for each row to
measure from. Finding one of these points is done in the following way:

\begin{equation}
    P = min(\sqrt{(x_i - \mu)^2})
    \label{eq:P}
\end{equation}

Where P is the value of the point, $x_i$ indicates the signal of one row, $\mu$ is the average depth of the points found using the derivative. This P has the value of the point and what is required is the index, so this can be found by simply searching for the point that has this value. This method is applied to one half of the image and then to the second half of the neck image, splitting the image in the horizontal axis. This results in two index points for each row, these are shown in the results section.

Once the points have been found the length from point to point for each row can be calculated. This is done using Pythagoras formula, and is done in the following way:

\begin{equation}
y = \sum_{n=0}^{N-1}\sqrt{1+|x[n]-x[n+1]|^2}
\label{eq:45}
\end{equation}

Here $y$ contains the length from point to point, $x[n]$ is the signal of one row, limited from the
first minimum point to the last minimum point calculated using equation \ref{eq:P}. This is done for each row and corresponds to half of the circumference of the neck, this can be seen in the result section.

\subsection{Reconstruction of full 3D data}
\begin{figure}
  \centering
  \includegraphics[scale=0.5]{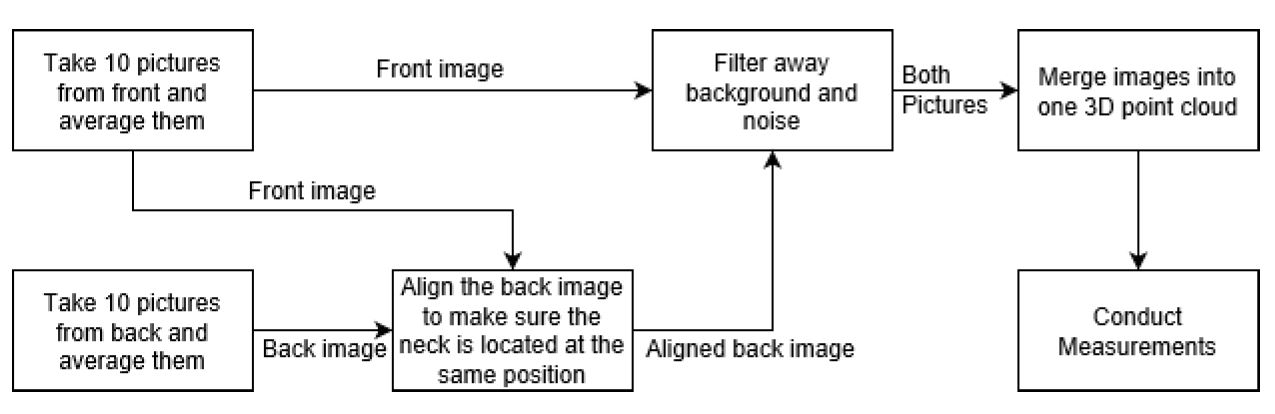}
  \caption{Flowchart of the second method using Kinect depth images from two different angles}
  \label{fig:fig4}
\end{figure}

In this section, a method of constructing the point cloud will be explained using two different
angles with reference to the human body. After that, measurements can be conducted on the
point cloud. The flow of this method can be seen in the flowchart shown in \ref{fig:fig4}.

\subsubsection{Positioning the patient}
\begin{figure}
  \centering
  \includegraphics[scale=0.4]{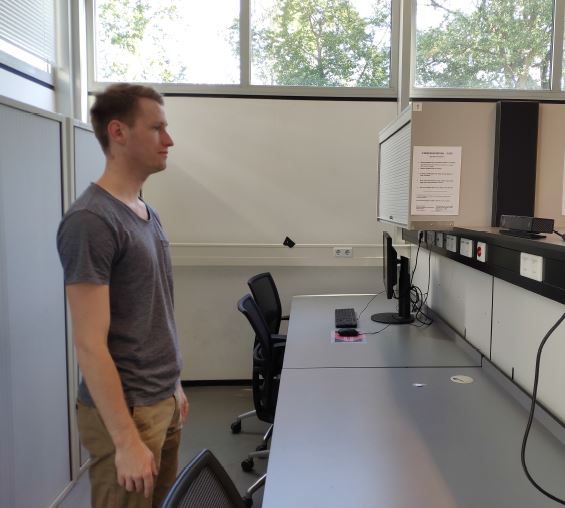}
  \caption{Camera set up and patient location}
  \label{fig:fig3}
\end{figure}

For the first image the patient is facing the camera directly, and the second image the patient is
rotated 180 degrees. The patient has to be located in both situations (front and back) at exactly
the same position except rotated 180 degrees. This was done by using a rotating platform and
rotate the platform by 180 degrees with the patient standing on the platform. The platform is
located one meter away from the camera similar to the one angle point cloud. The method is
the same as in the one angle point cloud (figure \ref{fig:fig3}) except for a rotating platform.

\subsubsection{Combining multiple data acquisition}
The next step is taking the pictures, this can be realized using the Image Acquisition Toolbox
from MATLAB (2019). Instead of taking one picture from one of the angles, multiple will be
taken from one of the angles. The number of pictures taken is 10, with the frame rate of the
Kinect of 30 Hz this will take one-third of a second. These 10 pictures will be combined into
one by taking the average of each pixel. This will cause some noisy points at the edges of the
person, but these will be filtered out. Combining the images is done both on the front and the
back images.

\subsubsection{Matching multiple data acquisition}
After combining images two images are the results, the front, and the back. These images contain the patient and the background, the background has to be removed. We know that the
patient is located one meter away from the camera, which will make the filtering the background out a lot more simple. Assuming that the patient’s head and neck are not larger than one meter in diameter, everything that has a depth value of larger than 1.5 meters and smaller than 0.5 meters can become zeros. For the matching part, it is better if these numbers are zero since the normalized correlation coefficient will give a better result. The back image will be shifted over the front image to find where they have the highest correlation. The back image is
shifted according to where the highest correlation is.

\subsubsection{Filtering point clouds}
In the matching part, the background has already been replaced by zeros, but in order to transform
the image into a point cloud, these have to be removed. This can simply be done by
detecting which are zero and deleting them. This already results in a decent point cloud of the patient from one angle, but there is still noise from merging the multiple images and from measurement error. This can be filtered by detecting and removing outliers, an outlier is detected
by first calculating the standard deviation from the mean of the average distance to neighbors
of all the points, second is analyze if its value is above a specified threshold. The neighbors are
the k-nearest points where k can be varied. In MATLAB (2019) this can be realized by using
the function pcdenoise(), which applies this to a point cloud object. These two parameters,
number of neighbors and threshold, can be adjusted until all the noise has been removed. We
have to make sure that the filter did not remove points that are necessary for doing the measurement.
Now the same has to be done for the back image to create a point cloud of the back.

\subsubsection{Merging point clouds}

Now there are two point cloud, one from the front and one from the back, and these have to be merged. The point cloud merging is done with the following steps:

• Equalize the means of both the point clouds, this can be realized by subtracting the difference in mean of the point clouds from the back point cloud. The reason for doing this is that there is no way for us to know what the exact center is of the patient. A reference point is necessary to merge the point clouds, and also a link between the point cloud and the reference should be know. The mean as a reference for the point clouds is very stable since it removes most variations.

• Rotate the back point cloud 180 degrees, this can be realized using a transformation matrix that rotates around the y-axis, which is the vertical axis. The matrix to rotate around the y-axis can be seen below:

\begin{equation}
R_y(\theta) = 
\begin{bmatrix}
cos(\theta) & 0 & sin(\theta) \\
0 & 1 & 0 \\
-sin(\theta) & 0 & cos(\theta)
\end{bmatrix}
\end{equation}

where theta is the angle, which is 180 degree in this case. Because of the rotation around
the y axis the depth values are negative. A simple absolute of the depth values will fix this.

• Combine the two point clouds, the means are the same for both point clouds which
means they are overlapping. To fix this a fixed distance between the two point clouds are created. This distance is a parameter that can be set at the beginning of the code and alters the resulting volume quite a lot (because you are changing the spacing between the two point clouds). This is not a problem as long as it is a constant distance for all future measurements of the patient. This procedure results in a nicely formed point cloud from two angles, this can be seen in the result section.

\subsubsection{Measuring the neck volume}
After creating a 3D point cloud the volume can be measured. This is done by scanning from top to bottom over the vertical axis and collecting all the areas of each slice. The area of each slice can be measured using the following formula:

\begin{equation}
A = (1/2)\sum_{n=0}^{N-2}(x_n+x_{n+1})(y_n-y_{n+1})
\end{equation}

Then the complete volume can be calculate as follows:

\begin{equation}
    V = \sum^{N}_{n=0}A_n\delta y
\end{equation}

where $A_n$ is the area of each slice, and $\delta$ y the height of each slice. But the complete volume is not what we need, only the volume of the neck is required. That is why a start point and an endpoint is required. How this is done is explained in the results section, and so is the results of this method.

\section{Experimental Results}
In this section, results of the two methods will be illustrated. First, the method of using one image to measure half of the circumference of the neck. Second, the method of using two different angles and merging the images into one to calculate the volume of the neck.

\subsection{Experimental results using one point cloud acquisition}
The results of the first method of using one image from one angle will be demonstrated here.

\subsubsection{Reference depth image}
\begin{figure}
  \centering
  \includegraphics[scale=0.5]{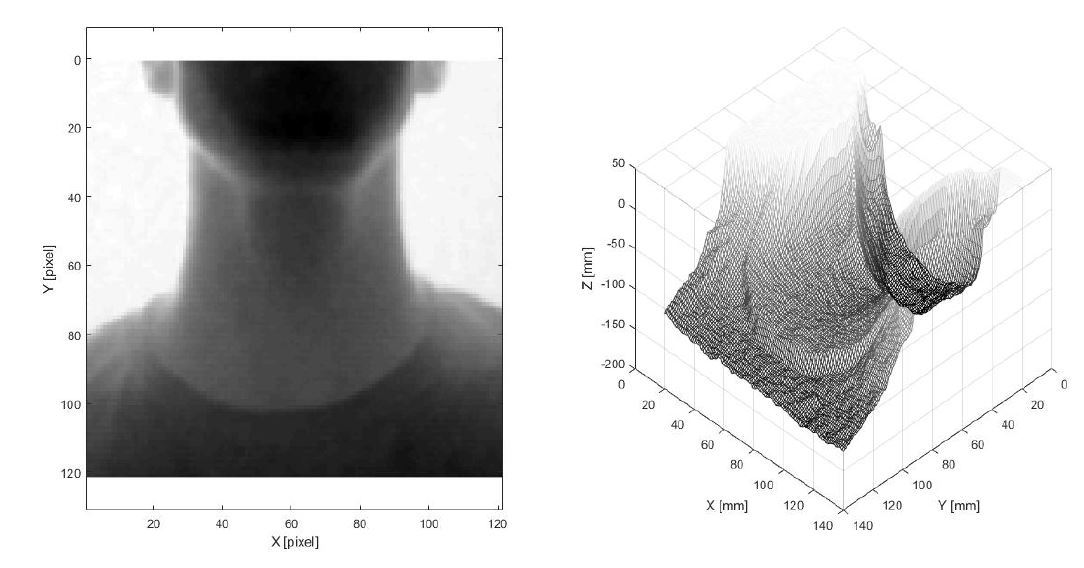}
  \caption{The reference neck used for finding center of the neck, left is the 2D version and right is a mesh of the same depth image in 3D}
  \label{fig:fig5}
\end{figure}

As mentioned before, the depth image focused on the neck will be used as a reference for locating
the center of the neck in future check-up sessions. The reference image can be seen in
figure \ref{fig:fig5}.

\subsubsection{Measurement points for circumference}
The points that will be used to measure half the circumference of the neck can be seen in figure \ref{fig:fig52}.

It can be seen that the points align all around the same depth. They do not exactly align because there are no points closer to the average depth. It can be improved by using linear interpolation between the points larger and smaller than the average depth. This will result in points all
exactly at the same depth. This was not yet implemented due to the change of focus to the 3D point cloud. It can also be seen that after y $\approx$ 80 the neck has ended. This value can also be
found by inspecting the derivative since after y $\approx$  80 the peaks will have a smaller value. So by determining when the peaks of the derivative cross a certain threshold, this point can be found. For the resulting circumference, this point is used as the endpoint.

\subsubsection{Resulting circumference}
\begin{figure}
  \centering
  \includegraphics[scale=0.4]{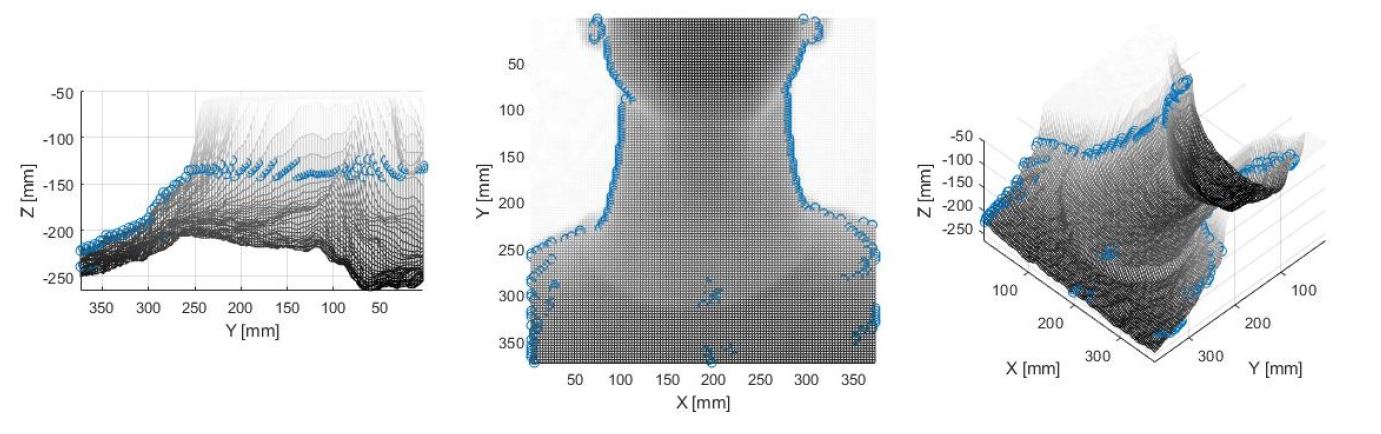}
  \caption{The points that are used to measure half the circumference of the neck from different angles}
  \label{fig:fig52}
\end{figure}

\begin{figure}
  \centering
  \includegraphics[scale=0.3]{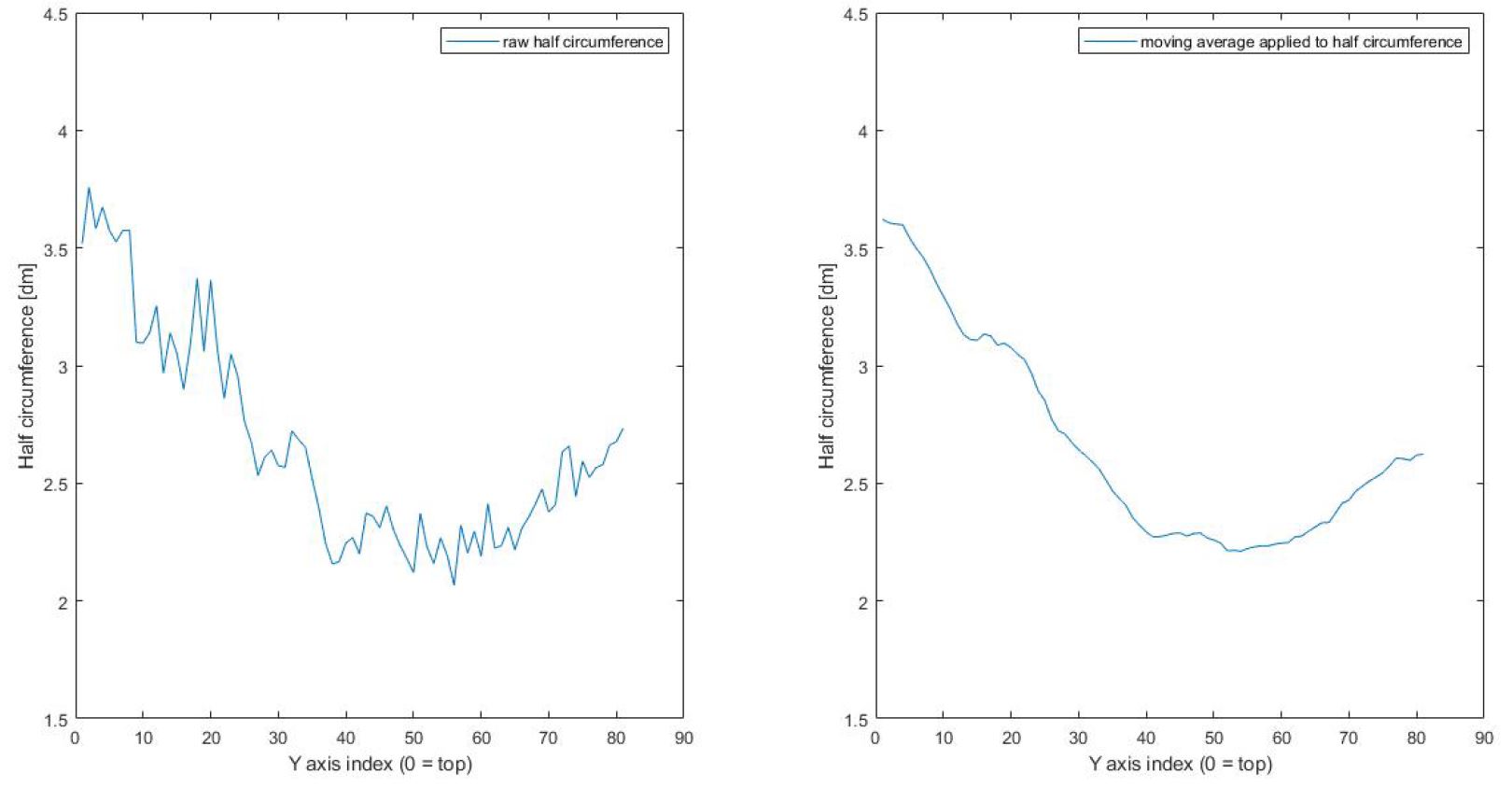}
  \caption{The raw circumference (left) and a moving average applied to the raw circumference (right)}
  \label{fig:fig6}
\end{figure}

After this the half the circumference has been calculated from point to point shown in figure \ref{fig:fig52}. This was done using equation \ref{eq:45} and the result can be seen in figure \ref{fig:fig6}.

It can be seen that the raw data still contains a lot of noise. But after applying a moving average over the signal the noise has been mostly removed. The noise in the raw circumference is due
to the fact that all the measurement points are not exactly at the same depth. This can be
improved by using linear interpolation between two points, where the first point is smaller
than the average depth value and the second larger. The right image of figure \ref{fig:fig6} can be used
to store as reference for the next session. By overlapping this reference and the next session’s
result anomalies can be detected. This has not been done in this study since our focus changed
to obtain the volume of the neck using a full 3D point cloud.

\subsection{Further improvement needs of the 3D point cloud} 
After getting feedback from medical doctors about our pre-results above, who preferred to see results based on the total volume of the
person, the model to conduct measurements on will be created using two different pictures from different angles. This will create a full 3D point cloud. In the following sub-section, it is shown how it is done.

\subsubsection{Before and after filtering}
\begin{figure}
  \centering
  \includegraphics[scale=0.3]{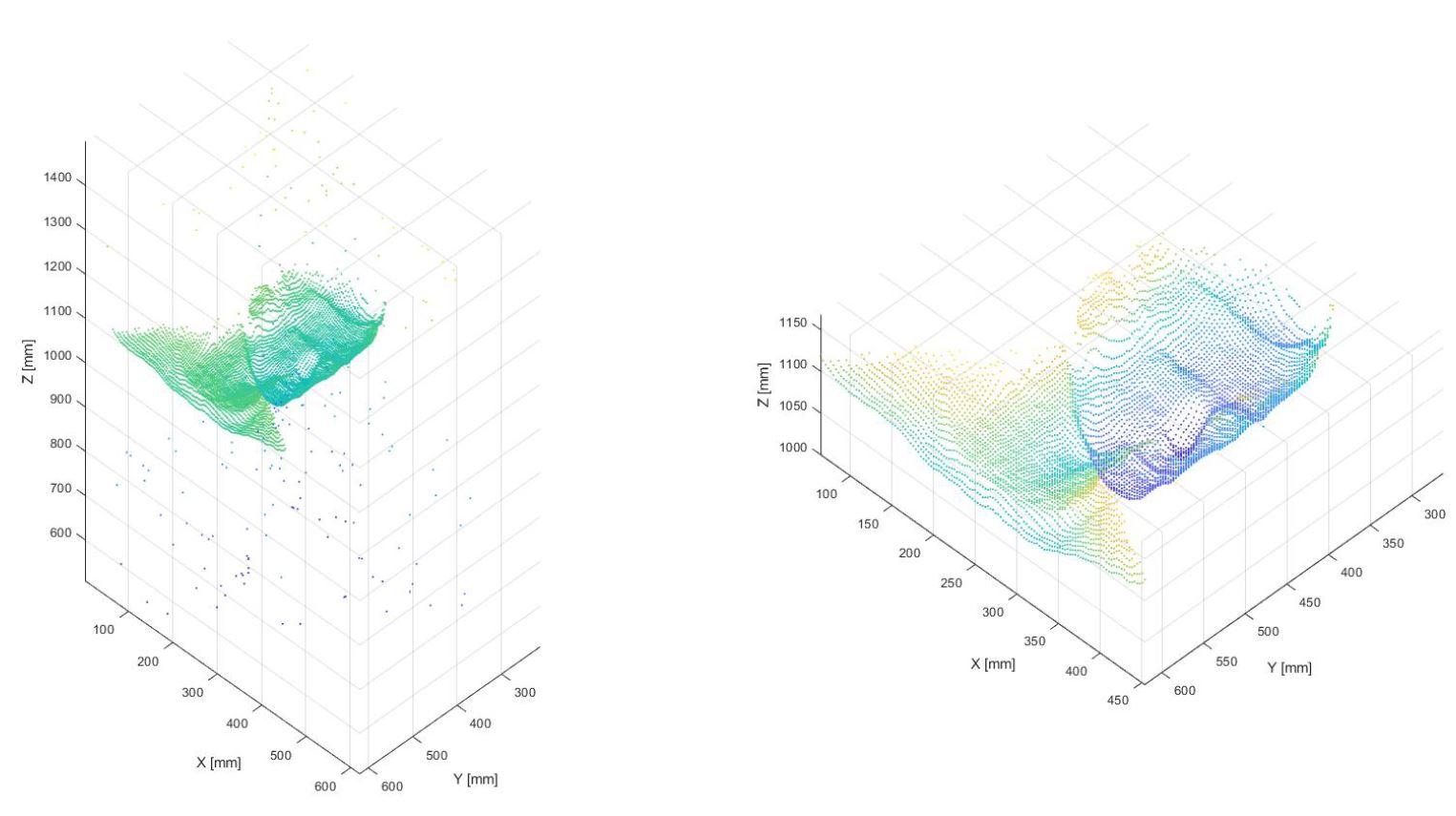}
  \caption{
  Before filter (left) and after filtering (right)}
  \label{fig:fig7}
\end{figure}

First, ten pictures from one angle will be taken and merged together (explained in the design).
After ten pictures have been taken and merged together, some noise remains, this is removed
by filtering. The filtering includes removing background and removing noise, removing the
background is simple so it is not shown here. The filtering of the noise is applied separately on the front and the back image, in figure \ref{fig:fig7} the filtering result of the front point cloud can be
seen.

\subsubsection{Full 3D point cloud of the patient} 
The filtering is done and applied to the front and back point cloud. The two images are aligned
with each other as explained previously. The next step is merging the two point clouds together
to make a full 3D point cloud. This can be seen in figure \ref{fig:fig8}.

\subsubsection{Area of the neck} 
From the full 360 point cloud of the patient, the area of each layer from top to bottom can be
measured. The result of this can be seen in \ref{fig:fig9}.

In the figure, the local minimum corresponds to the neck area since this is the smallest area of
the patient when it is evaluated from top to bottom. It can be seen that at this local minimum
the area of all measurements are overlapping quite accurately. In order to calculate the volume
of the neck only, a starting point and an endpoint are required. This start point and endpoint
is based on the local minimum of the neck. Once this point is found a value based on the
prominence of the peak will be used to find the start and endpoint, the best way to show this is
in \ref{fig:fig10}.

This is done for each of the signals in figure \ref{fig:fig9}, and this results in figure \ref{fig:fig11}.

This method shifts the areas according to the minimum point and overlaps them. The last step is multiplying each value by $\delta y$ and sum these values together. This results in the volume of the neck in $dm^3$ which is the same as liters.

\subsubsection{Accuracy and Precision} 
Precision can be determined by statistical variability, and accuracy can be determined by measuring
how close the result is to the true value. To test these properties of the system the following
test and calculations will be conducted:

• 10 measurements will be taken of the patient without anything attached to the neck

• 10 measurements will be taken of the patient with something attached to the neck. The
thing that is attached to the neck will be clay, where the volume is measured beforehand.

• The volume will be calculated for all these measurements. After that, the standard variability
will be calculated, which will give an indication of the precision.

The difference in the average volume of the first 10 measurements and the second 10
measurements will give an indication of the accuracy.
The clay that was used has a simple shape that can be attached to the neck. The clay can be
seen in figure \ref{fig:fig12}.

The method of measuring the volume of the clay is by submerging it into a measuring cup. This
can be seen in figure \ref{fig:fig13}.

With these two images, the difference in the water level is equal to the volume, which is approximately
60 ml. The result of the method explained at the beginning of this subsection can be seen in the list given in figure \ref{fig:table51}
below.

\section{Further reflections on the experiments}

The aim of the experiment was to compare two different methods of getting volume information
about the neck. One was using a depth image from one point of view, and the second using
two depth images from two points of view. The second aim was to see how accurate and precise
the method is compared to real life. To do this a piece of clay with known volume was used
and attached to the neck. Afterward, the volume of the neck with the clay was measured and
compared with a previous measurement without the clay.
From figure \ref{fig:fig6} it can be seen that the raw data from using a single image is very noisy. This
noise is due to not using linear interpolation between the points. This method has not been
thoroughly tested to evaluate if it can detect a change in the circumference of the neck. From
figure \ref{fig:fig11} it can be seen that this method using two images contains a lot less noise compared
to the method using a single image.
From the list at \ref{fig:table52} it can be seen that the accuracy is slightly off by 10.5$\%$ compared to the real volume.
This might be due to some wrong estimations in the system, and also due to the Kinect
not seeing everything. The estimation that was made is that each pixel corresponds to a certain
amount of millimeters in real life. This was converted using a constant factor estimated beforehand,
it is possible that this factor is slightly deviated from what it should be. The second
thing is that the clay used was not completely flat, so when a picture was taken by the Kinect it
is likely that some of the clay’s shape was changed due to shadowing in the depth image. This
can only increase the volume of the clay just like what the results show since it shows 6.3 ml
larger than it should be.
The precision of the system was determined using the standard deviation. The standard deviation
of the system is 42.4 ml, which is 17.6 ml smaller than the real volume of the clay. This
is problematic if only one measurement would be taken since it is possible that the doctor will
neglect the status of the lymphedema or over-exaggerates the status of the lymphedema. That
is why the suggestion is to take multiple measurements for each session and estimate the average.
This will give a better indication of the status due to the accuracy of the system.

\section{Conclusions}

The current method of using a measuring tape to detect the status of lymphedema in the neck
is inconsistent and is prone to human error. The method of using two images from different angles
has an error of 10.5$\%$ compared to the real volume when taking multiple measurements.
The first research question in the introduction is ’How accurate and precise is this method for
detecting a change in volume in the neck area caused by lymphedema?’. This method is able to
detect if the status of the patient is improving or declining since this directly relates to volume
increasing/decreasing. There is no real measurable quantity that exists at the moment to determine
the status of the neck lymphedema, that is why it is impossible to say when the volume
increase/decrease is problematic. The doctor can determine if something is problematic, but
hopefully, the change in volume in the neck is a good indicator.
The second research question is ’How many pictures are required from different angles to create
a model that can be used for detecting lymphedema?’. A model created from one angle caused
a lot of noise to be on the resulting circumference. However, a model created from two angles
caused good results, but can be improved, since the point cloud still contains a lot of points
missing at the neck area. A point cloud created from three angles will result in a solid point
cloud.

\begin{figure}
  \centering
  \includegraphics[scale=0.4]{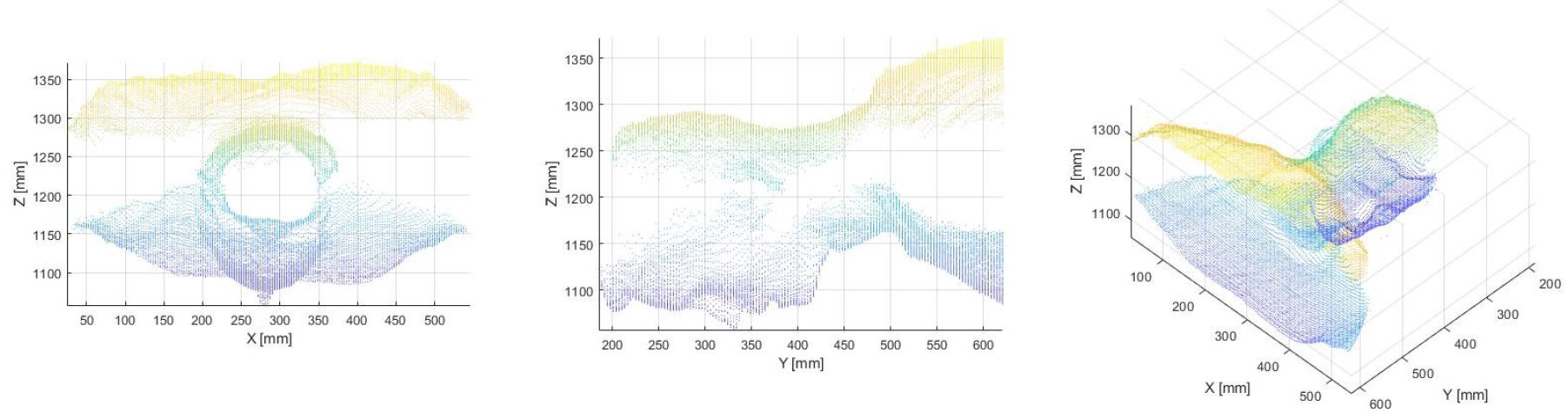}
  \caption{A complete 3D point cloud of the patient viewed from three different angles}
  \label{fig:fig8}
\end{figure}

\begin{figure}
  \centering
  \includegraphics[scale=0.4]{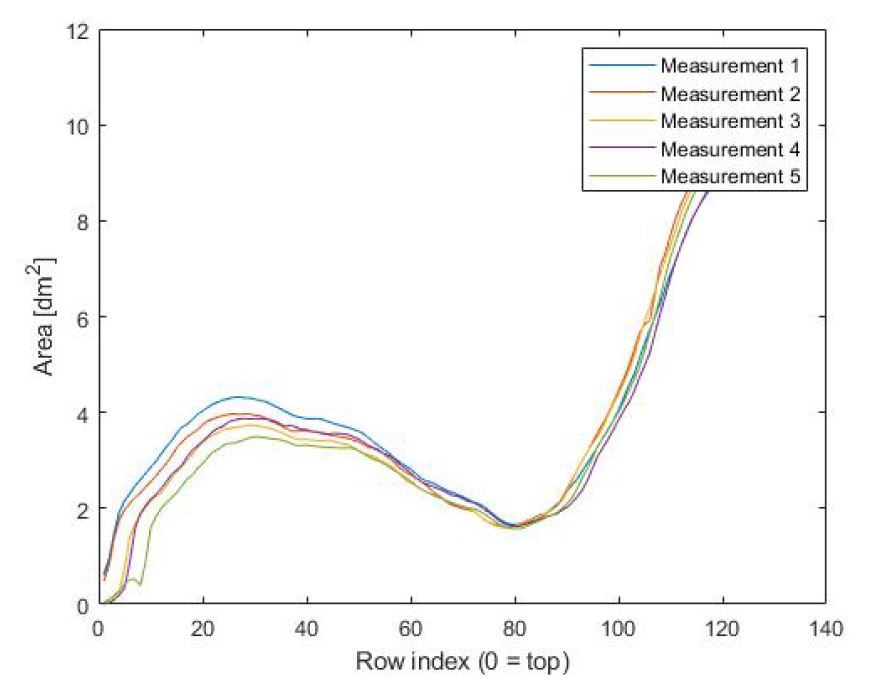}
  \caption{Area of the total point cloud of the patient from top to bottom}
  \label{fig:fig9}
\end{figure}

\begin{figure}
  \centering
  \includegraphics[scale=0.5]{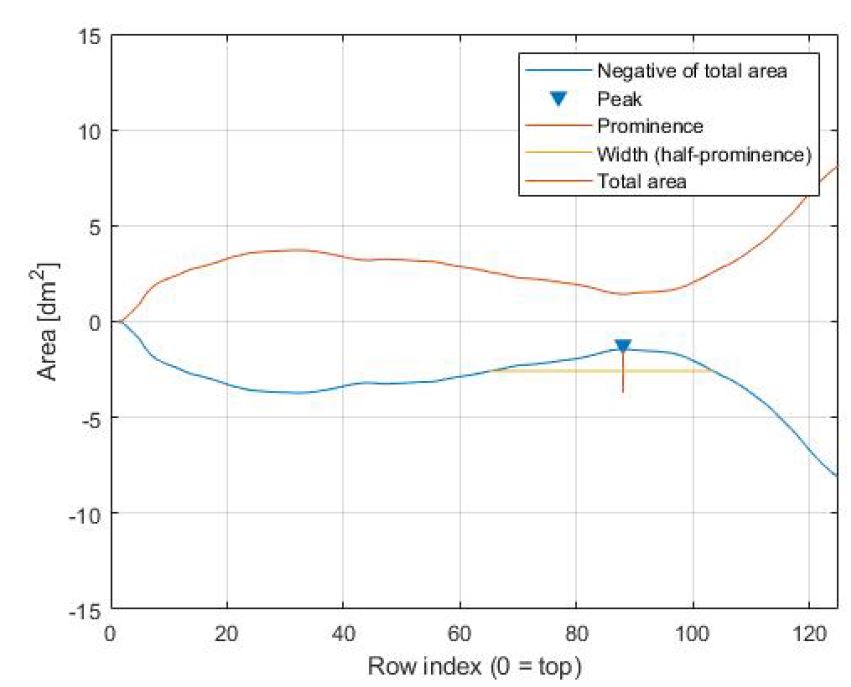}
  \caption{Displaying the method of narrowing down the neck area to calculate the volume}
  \label{fig:fig10}
\end{figure}

\begin{figure}
  \centering
  \includegraphics[scale=0.5]{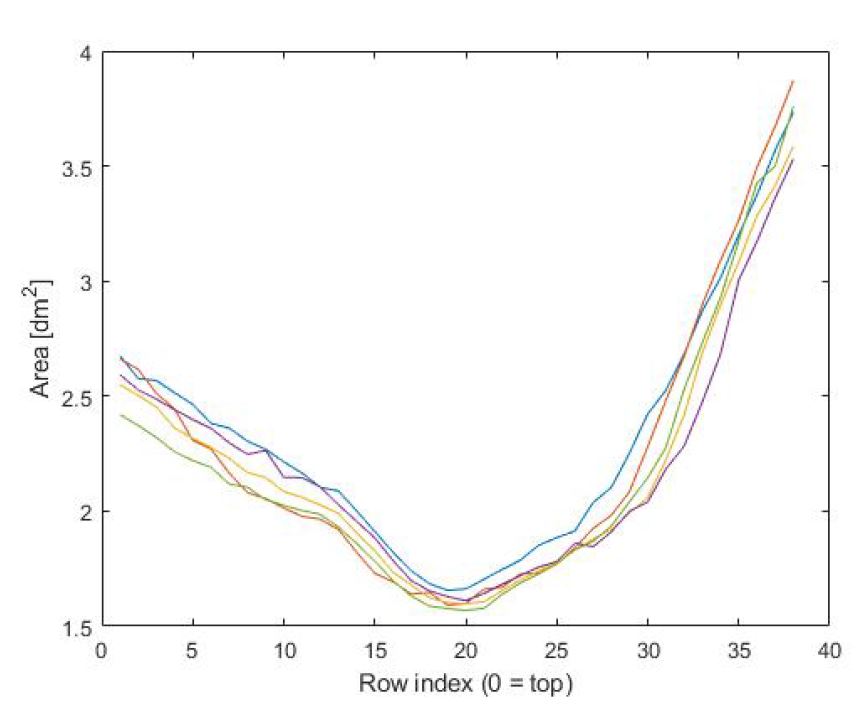}
  \caption{Narrowed down neck}
  \label{fig:fig11}
\end{figure}

\begin{figure}
  \centering
  \includegraphics[scale=0.5]{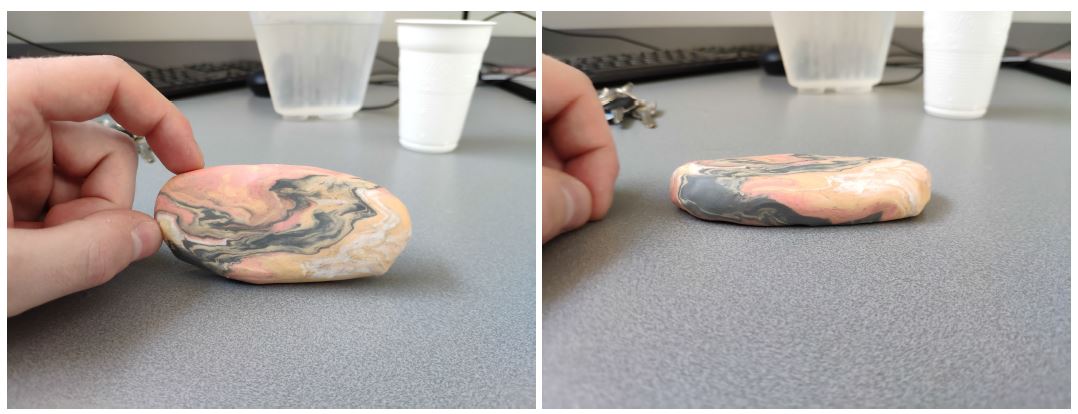}
  \caption{The shape of the clay from the top and the side}
  \label{fig:fig12}
\end{figure}

\begin{figure}
  \centering
  \includegraphics[scale=0.3]{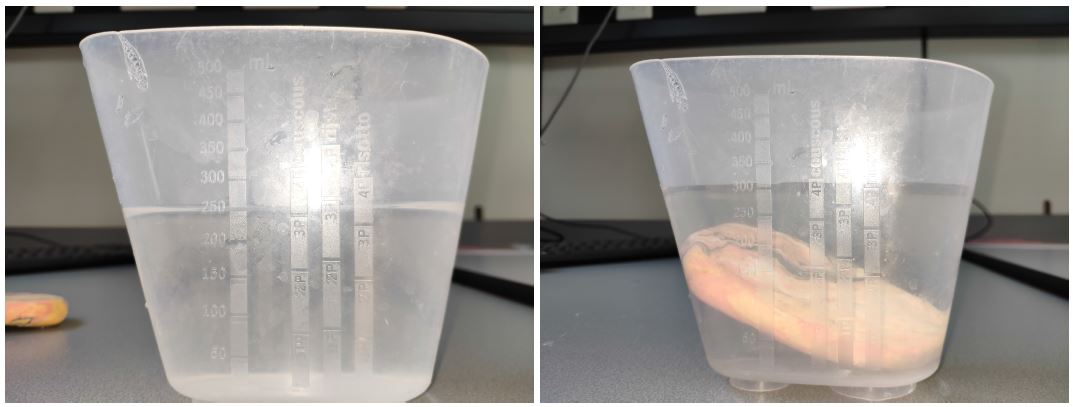}
  \caption{The water before submerging (top), the water after submerging (bottom)}
  \label{fig:fig13}
\end{figure}

What should be improved in a future version of this method is a couple of things listed below:

• The setup uses two angles to construct a 3D point cloud. This can be increased to three since it can be seen in figure \ref{fig:fig8} that the point cloud has a gap in the middle which contains no points.

• The constant used to convert pixels into millimeters should be calculated from the image itself. This can be done by getting a reference object in the depth image where the area is known of.

• The method used for finding a feature based on the minimum area of a layer from the neck should be improved. The minimum area of a layer from the neck can shift due to swelling caused by lymphedema. This can cause the problem that a wrong comparison can occur between two check-up sessions. A better method would be by finding a fixed reference point that does not change due to swelling in the neck. For example, the nose or lower lip can be used for this.

\begin{figure}
  \centering
  \includegraphics[scale=0.5]{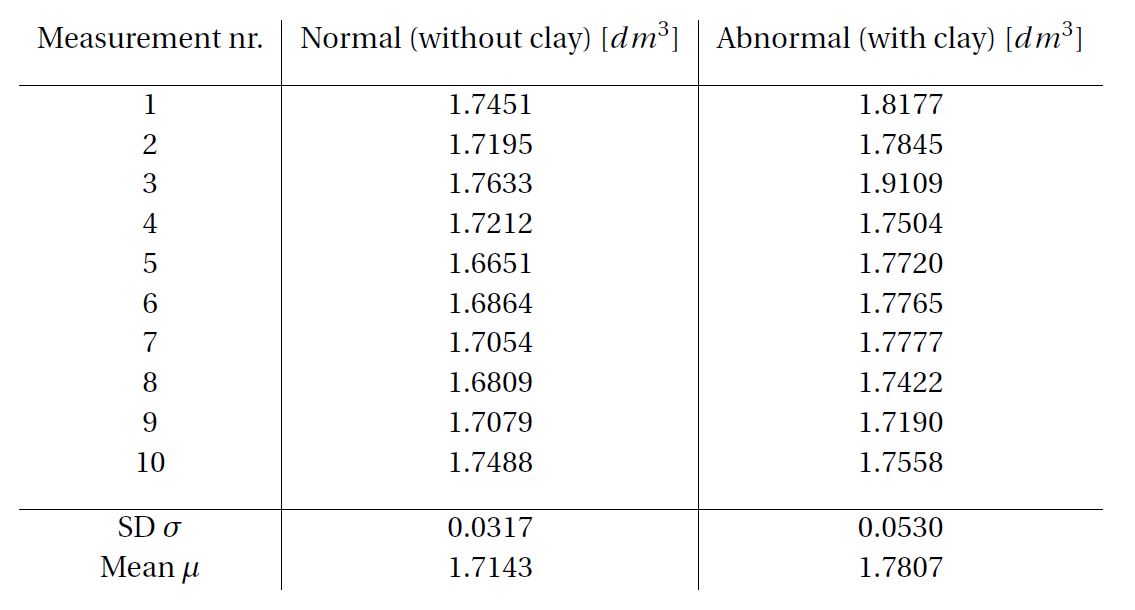}
  \caption{Table containing the measurements}
  \label{fig:table51}
\end{figure}
 
\begin{figure}
  \centering
  \includegraphics[scale=0.5]{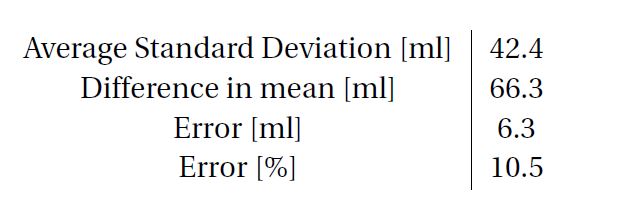}
  \caption{Statistics of the measurements}
  \label{fig:table52}
\end{figure}

\bibliographystyle{unsrt}  
\bibliography{main}  %%% Remove comment to use the external .bib file (using bibtex).

\end{document}